\documentclass[conference]{./support/ieeeconf}
\usepackage{times}
\usepackage{amsmath}
\usepackage[pdftex]{graphicx}
\usepackage{caption}
\usepackage{subcaption}
\usepackage{amsmath,amssymb,amsopn,amstext,amsfonts}
\usepackage{cancel}
\usepackage[space]{cite}
\usepackage{pdfsync}
\usepackage{balance}
\usepackage{color}
\usepackage{algorithm}
\usepackage{algorithmic}
\usepackage{url}
\usepackage{booktabs}
\usepackage{threeparttable}
\usepackage[linkcolor=black,citecolor=black,urlcolor=black,colorlinks=true]{hyperref}
\usepackage{multirow}
\usepackage{epstopdf}
\usepackage{ulem}
\usepackage{graphicx}
\usepackage{array}
\usepackage{verbatim}
\usepackage{makecell}
\usepackage{array}
\usepackage{svg}

\IEEEoverridecommandlockouts

\newcommand{\revise}[1]{\textcolor{black}{#1}}

\graphicspath{{./figure/}}

\DeclareGraphicsExtensions{.pdf,.png,.jpg,.eps,.PNG}
\title{\LARGE \bf MapLocNet: Coarse-to-Fine Feature Registration for Visual Re-Localization in Navigation Maps}

\author{Hang Wu, Zhenghao Zhang, Siyuan Lin, Xiangru Mu,  Qiang Zhao, Ming Yang, Tong Qin$^*$
		\thanks{
  Tong Qin, Xiangru Mu, and Ming Yang are with Global Institute of Future Technology, Shanghai Jiao Tong University, Shanghai, China.
  Hang Wu, Zhenghao Zhang, Siyuan Lin, and Qiang Zhao are with IAS BU, Huawei Technologies, Shanghai, China.
		{\tt\small 
  \{qintong, muxiangru, mingyang\}@sjtu.edu.cn
  \{wuhang12, zhangzhenghao6, linsiyuan1, zhaoqiang20\}@huawei.com}. 
		{  $^*$ is the corresponding author}.
	}}

\begin{document}

\maketitle
\thispagestyle{empty}
\pagestyle{empty}

\begin{abstract}

Robust localization is the cornerstone of autonomous driving, especially in challenging urban environments where GPS signals suffer from multipath errors.
Traditional localization approaches rely on high-definition (HD) maps, which consist of precisely annotated landmarks. 
However, building HD map is expensive and challenging to scale up.
Given these limitations, leveraging navigation maps has emerged as a promising low-cost alternative for localization.
Current approaches based on navigation maps can achieve highly accurate localization, but their complex matching strategies lead to unacceptable inference latency that fails to meet the real-time demands.
To address these limitations, we propose a novel transformer-based neural re-localization method. Inspired by image registration, our approach performs a coarse-to-fine neural feature registration between navigation map and visual bird's-eye view features.
Our method significantly outperforms the current state-of-the-art OrienterNet on both the nuScenes and Argoverse datasets, which is nearly $10\%/20\%$ localization accuracy and $30/16$ FPS improvement on single-view and surround-view input settings, separately. 
We highlight that our research presents an HD-map-free localization method for autonomous driving, offering cost-effective, reliable, and scalable performance in challenging driving environments.

\end{abstract}

\section{Introduction}

With the recent development of autonomous driving over the past decade, robust localization plays a crucial role.
\revise{Both autonomous driving vehicles and human-driving navigation rely heavily on Global Navigation Satellite Systems (GNSS) for outdoor localization, but these signals are prone to be noisy in urban areas. 
Multipath propagation errors from surrounding infrastructure as well as line-of-sight occlusion from buildings, tunnels, bridges etc. can severely impact GPS localization accuracy.}
Without an effective global localization source, the location drifts quickly.

To survive in a GPS-denied situation, additional active localization methods are required.
By leveraging prior built maps, such as 3D point clouds and distinctive visual features, LiDAR-based \cite{zhang2014loam} and visual-based \cite{mur2015orb, qin2018vins} SLAM methods can be used for localization.
However, this point-wise prior map is memory-consuming, and can not be used in the large environment for the autonomous driving task.
\revise{Autonomous driving has relied heavily on high-definition (HD) maps containing precisely geo-referenced landmarks and geometries in the GPS-denied area}.
\revise{However, the exorbitant costs of producing and maintaining these maps have severely limited their scalability across diverse environments and geographies.}
\revise{As a result, the dependency on HD maps has been a major bottleneck preventing wider adoption of self-driving capabilities.}
\begin{figure}[t]
	\centering
	\includegraphics[width=0.45\textwidth]{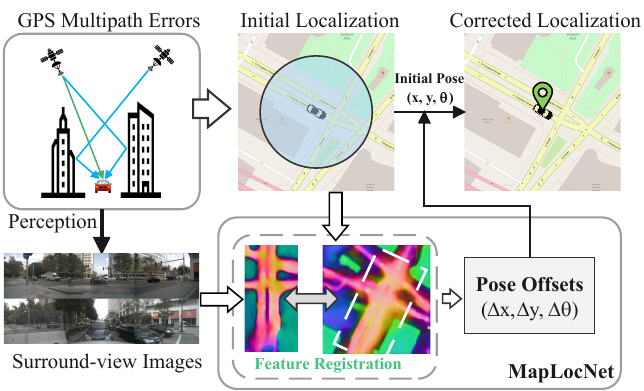}
	\caption{
 Due to signal occlusion and multipath errors, the GPS-based positioning is unreliable in complex urban environments.
 To address this problem, we propose MapLocNet, which leverages surround-view images and navigation maps, utilizing consecutive neural localization modules based on coarse-to-fine feature registration principles to achieve superior localization accuracy in challenging scenarios.}
	\label{introduction}
	\vspace{-0.5cm}
\end{figure}
\revise{With the advancement in perception algorithms, methods such as HDMapNet\cite{hdmapnet} and MapTR\cite{MapTR} have enabled online HD maps generation, allowing autonomous driving to be possible even with low-precision localization. }

On the other hand, we find some cues from biology, that human drivers can recognize location just with navigation maps.
By correlating visual observations with map information, humans can roughly localize themselves in complex urban environments.
Humans extract high-level semantics like road structures, building outlines, and landmarks from their surroundings, and leverage cognitive abilities to match these semantics with navigation maps.
Currently, in the fields of robotics and augmented reality (AR), similar approaches \cite{sarlin2023orienternet, sarlin2023snap} have been proposed to emulate human-like localization methods.
However, such methods often employ complex matching strategies for localization, making real-time inference unattainable.
As a result, they cannot be applied in autonomous driving systems.
To address the aforementioned challenges, we propose MapLocNet, a novel approach that achieves high localization accuracy while meeting real-time performance requirements. 
In our approach, we encode surround-view images into the BEV space and process the navigation map using a U-Net\cite{ronneberger2015u}. The key innovation is the use of a transformer-based hierarchical feature registration method, which aligns visual BEV features with map features effectively, resulting in highly accurate localization.
Our proposed approach surpasses current state-of-the-art (SOTA) methods in both localization accuracy and inference latency.
Overall, the contribution of our paper is summarized as follows:
\begin{itemize}
    \item We proposed MapLocNet, achieving highly accurate localization by fusing surround-view images and navigation maps, especially in GPS-denied areas suffered from significant positioning drift.
    \item We introduce a hierarchical coarse-to-fine feature registration strategy that aligns BEV and map features, attaining superior localization accuracy and inference speed compared to existing methods.
    \item We develop a novel training criterion that leverages perception tasks as auxiliary objectives for pose prediction. Our MapLocNet achieved the SOTA localization accuracy on the both nuScenes and Argoverse datasets.
\end{itemize}

We highlight that our research presents an HD-map-free, reliable and human-like localization approach, achieving superior localization accuracy compared to existing methods.
  % qin tong

\section{literature review}

\subsection{Localization Using \revise{Navigation} Map\revise{s}}

Building the HD map was expensive, recent research focused on localization based on the lightweight navigation map. 
Panphattarasap \textit{et al.} \cite{automated_map_reading} proposed a novel approach to image-based localization in urban environments using semantic matching between images and a 2D map. 
Samano \textit{et al.} \cite{you_are_here} designed a novel method to geo-localize panoramic images on a 2D navigation map based on learning a low dimensional embedded space.
Zhou \textit{et al.} \cite{image_based_geolocalization} presented a 2.5D map-based cross-view localization method that fuses the 2D image features and 2.5D maps to increase the distinctiveness of location embeddings.
OrienterNet \cite{sarlin2023orienternet} proposed a deep neural network that estimates the pose of a query image by matching a neural BEV with available maps from OpenStreetMap (OSM)\cite{haklay2008openstreetmap} and has achieved high-precision localization.
Other methods \cite{slicematch, uncertainty_aware_geolocalization, visual_cross_view} achieved cross-view geolocalization that matches the camera images from vehicles with an aerial image or a satellite image to determine the vehicle’s pose.
Drawing inspiration from the previous researches, we propose a localization approach that combines visual environmental perception with navigation maps.

\begin{figure*}[t]
	\centering
	\includegraphics[width=0.9\linewidth]{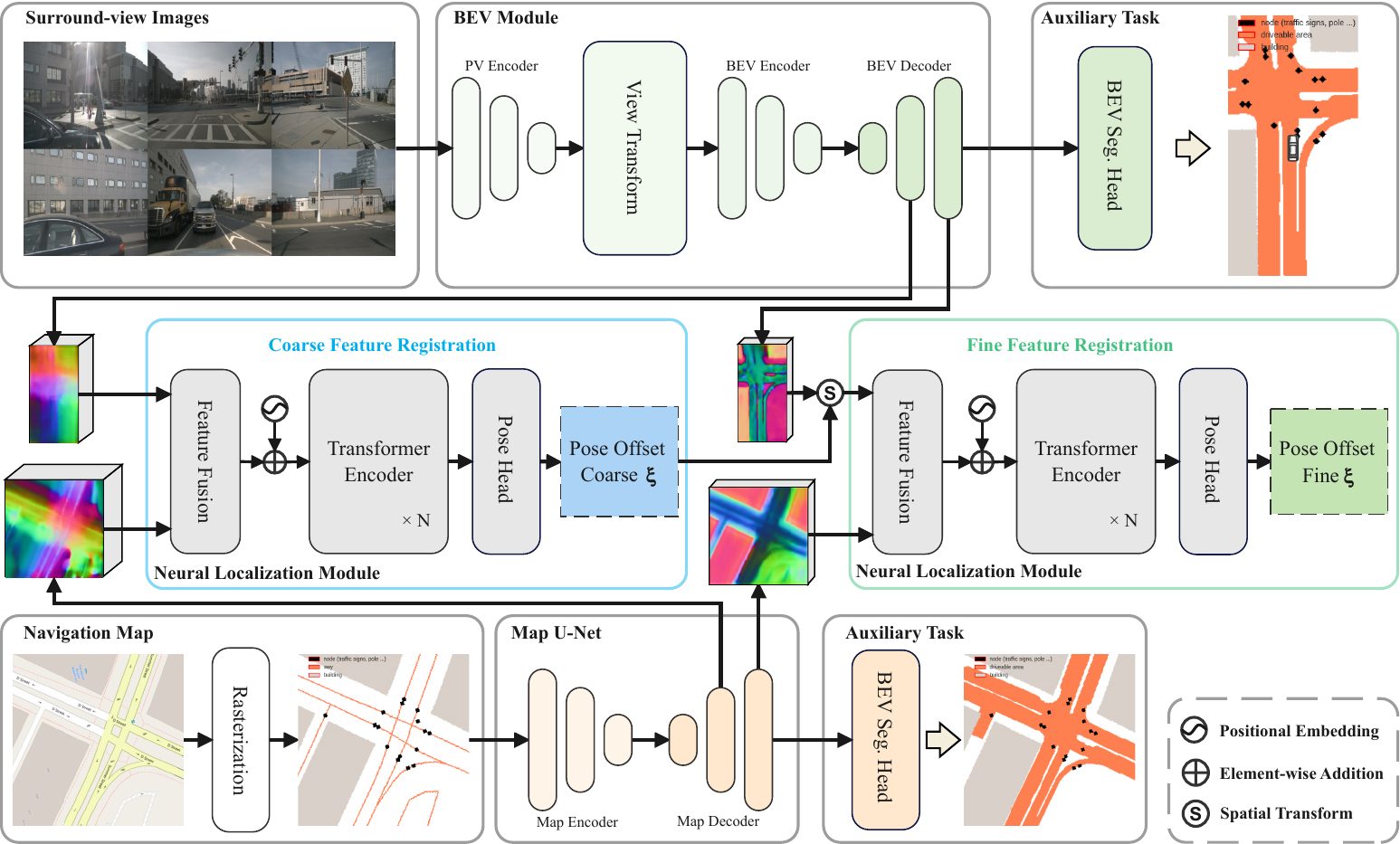}
	\caption{The overall architecture of MapLocNet comprises three main modules: the BEV Module, Map U-Net, and Neural Localization Module. Our approach employs a coarse-to-fine feature registration strategy, extracting multi-scale features from both the BEV Decoder and Map Decoder to perform hierarchical feature alignment. Following the initial coarse registration stage, which yields a coarse estimate of the pose offset, we apply a spatial transformation to the high-resolution BEV features to facilitate the subsequent fine registration process. The predictions from both stages are combined to yield the final pose offset estimation result.
 }
	\label{overview}
	%\vspace{-1.0cm}
\end{figure*}

\subsection{BEV Representation for Visual Localization}

There are many methods to transform the image features to the BEV grid, including geometry methods and learning-based methods.
Cam2BEV \cite{cam2bev} and VectorMapNet\cite{vectormapnet} used a geometry method that leveraged Inverse Perspective Mapping (IPM) to transform the image features to the BEV space through the plane assumption. 
HDMapNet \cite{hdmapnet} put forward a novel view transformer that consists of both neural feature extraction and geometric projection to get the BEV features.
LSS \cite{lss}, BEVDepth \cite{bevdepth}, BEVDet \cite{bevdet} learned a depth distribution of image features to lift each pixel to the 3D space.
Then they used the camera extrinsic and intrinsic to splat all frustums into the BEV.
GKT \cite{GKT} proposed an efficient and robust 2D-to-BEV representation learning method that leveraged the geometric priors to guide the transformer to focus on discriminative regions, and unfolded kernel features to obtain BEV features.
BEVFormer \cite{bevformer} leveraged predefined grid-shaped BEV queries to look up spatio-temporal space and aggregate spatio-temporal information from images, achieving SOTA performance on 3D object detection.
To balance the precision and efficiency, we designed our BEV module based on the LSS architecture. 

\subsection{Image Registration}
\revise{Image registration aims to find the spatial mapping between pixels in one image and another image, which is widely used in medical imaging and robotics research.}
\revise{Traditional feature-based methods\cite{sift, orb} leveraged the keypoints detected from images and its descriptors to match different images.
Recently, CNN and transformer based image registration methods emerged to accelerate the registration time and accuracy.}
DIRNet\cite{DIRNET} proposed a deep learning network for deformable image registration.
The network included a ConvNet regressor, a spatial transformer, and a resampler.
C2F-ViT\cite{mok2022affine} is a learning-based approach for 3D affine medical image registration that leverages the global connectivity of the self-attention mechanism and the locality of convolutional feed-forward layers to robustly encode global orientations and spatial relationships into a set of geometric transformation parameters.
Following the principles of C2F-ViT, we construct a hierarchical feature registration module for visual localization.

\subsection{End-to-end Localization Neural Networks}

Recent end-to-end (E2E) approaches have introduced efficient architectures that estimate ego pose directly from sensor inputs and prior maps, circumventing the need for complex geometric calculations and hand-crafted rules.
Using a differentiable optimization method, PixLoc \cite{PixLoc} designed an E2E neural network to estimate the pose of an image by aligning deep features with a reference 3D model.
I2D-Loc \cite{I2D-Loc} proposed an effective network for camera localization based on the local image-LiDAR depth registration and used the BPnP module to calculate the gradients of the backend pose estimation for E2E training.
BEV-Locator \cite{bev-locator} designed a novel E2E architecture for visual semantic localization from multi-view images and a vectorized global map.
Based on the cross-modal transformer structure, it addressed the key challenge of cross-modality matching between semantic map elements and camera images.
EgoVM \cite{egovm} built an E2E localization network that used light vectorized maps and has achieved centimeter-level localization accuracy.
% The method adopted learnable semantic embeddings and a transformer decoder to bridge the gap between vectorized maps and BEV features.
Inspired by the aforementioned works, our method builds an E2E localization network based on the transformer to achieve accurate localization.
 % xiangru

\section{methodology} %wuhang

\subsection{Problem Formulation and System Overview}

Given an initial vehicle localization $\mathbf{\check{p}} = (x, y,\theta)$ under noisy GPS, our goal is to estimate the transformation $\Delta \mathbf{\hat{T}} = \{\mathbf{R}, \mathbf{t}\}$ that transforms the initial noisy position to the ground-truth position $\mathbf{p}$. Since the re-localization is performed on a 2D navigation map, the %typical 6-DoF 
pose transformation can be simplified to a 3-DoF transformation with a %2D 
rotation $\mathbf{R} \in SO(2)$ and a 2D translation $\mathbf{t} \in \mathbb{R}^2$.
The transformation can be solved by:
\begin{equation}
	\underset{\Delta \mathbf{\hat{T}}}{min}\left\|\Delta \mathbf{\hat{T}}\cdot \mathbf{\check{p}}, \mathbf{p} \right\|_{2}.
    \label{eq:1}
\end{equation}
Here, $\Delta \mathbf{\hat{T}}$ represents the transformation $\mathbf{\hat{T}_{est\leftarrow err}}$ from the erroneous localization to the estimated localization. We aim to minimize the difference between $\mathbf{\hat{T}_{est\leftarrow err}}$ and the ground-truth transformation $\mathbf{T_{GT\leftarrow err}}$. 
% Since we are dealing with a 3-DoF problem, 
$\Delta \mathbf{\hat{T}}$ is denoted as $\boldsymbol{\hat{\xi}} = \left ( \Delta x,\Delta y,\Delta \theta  \right )$, where $\left (\Delta x,\Delta y \right )\in \mathbb{R}^2$ represent the longitudinal and lateral offsets, and $\Delta \theta \in \left ( -\pi ,\pi  \right ]$ represents the heading angle offset.

Our localization approach contains three key \revise{modules}: the BEV \revise{Module}, the Map \revise{U-Net} and the \revise{Neural} Localization \revise{Module}. The overview of our network are shown in Fig. \ref{overview}.
The entire network is trained end-to-end using surround-view images, navigation maps, BEV segmentation, and ground-truth pose offsets. We leverage coarse and fine pose offsets regression and semantic segmentation losses during training to supervise consistent representations among visual BEV and map branchs. Details of the architecture and losses will be presented in the following sections.

\begin{figure}[t]
    % \centering
    % \begin{subfigure}{0.9\linewidth}
    %     \centering
    %     \includegraphics[height=2.8cm]{figures/camera.jpg}
    %     \caption{Surround-view Images}
    %     \label{fig:sub3}
    % \end{subfigure}

    % \vspace{2mm}

	\centering
	\begin{subfigure}[b]{0.40\linewidth}
		\centering
		\includegraphics[height=3.3cm]{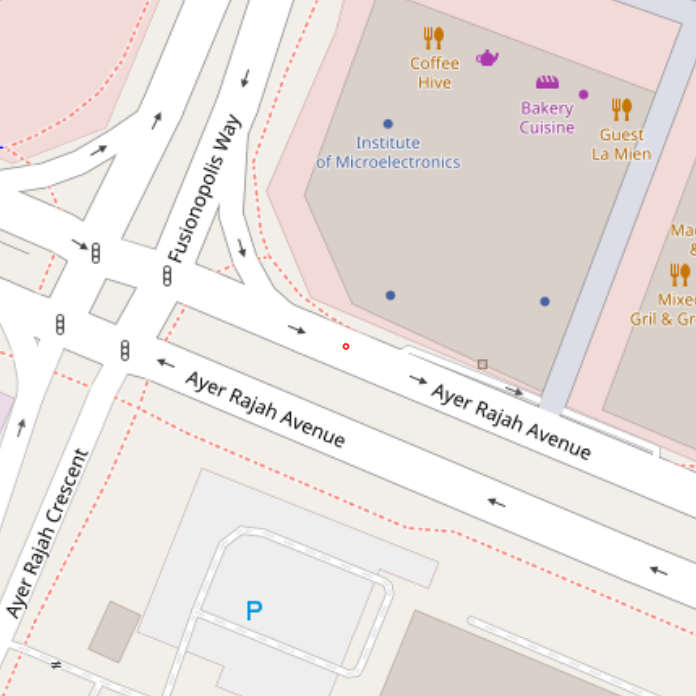}
		\caption{Original Map}
		\label{fig:left}
	\end{subfigure}
    \hspace{0.5cm}
	\begin{subfigure}[b]{0.40\linewidth}
		\centering
		\includegraphics[height=3.3cm]{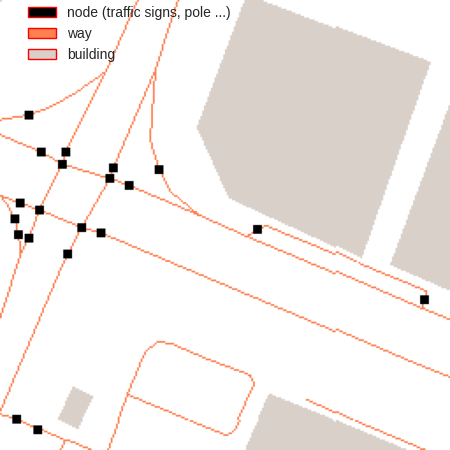}
		\caption{Rasterized Map}
		\label{fig:right}
	\end{subfigure}
 
    \vspace{2mm}
    
    \begin{subfigure}{0.9\linewidth}
        \centering
        \includegraphics[height=2.8cm]{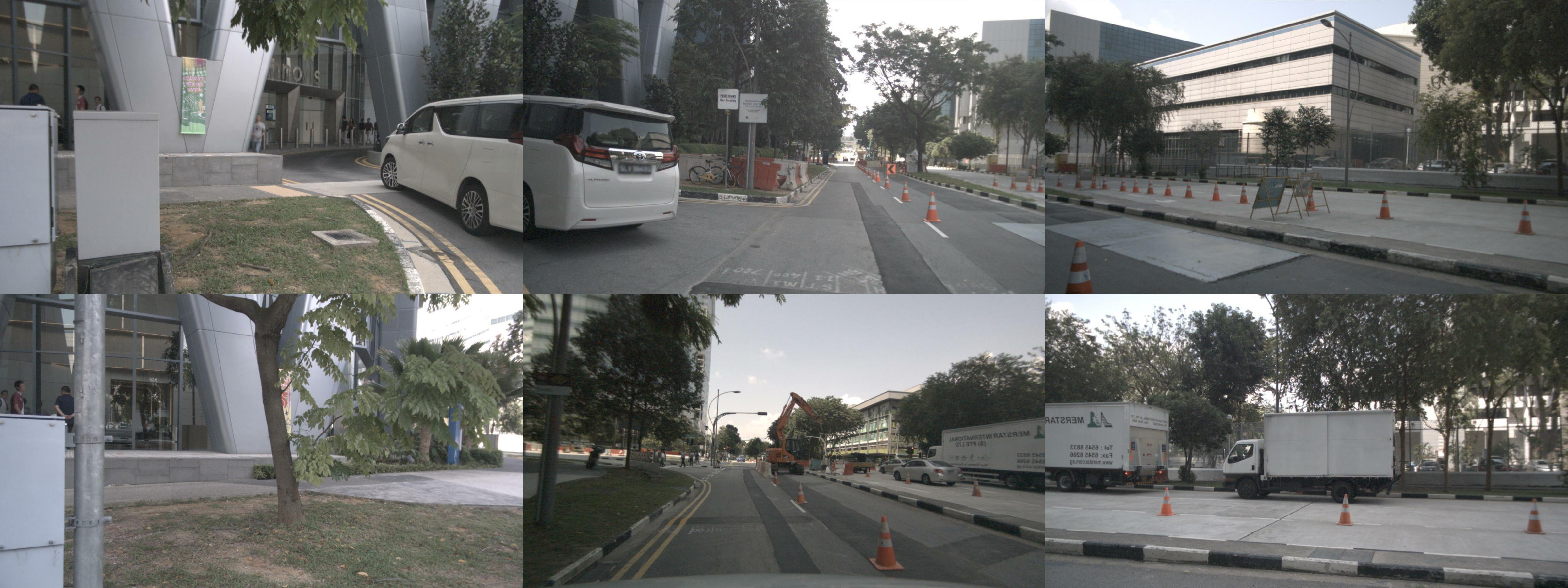}
        \caption{Surround-view Images}
        \label{fig:sub3}
    \end{subfigure}
    
	\caption{\revise{Visualization of the original navigation map, its rasterized representation, and corresponding surround-view images. Rasterization enhances the expression of topological cues and spatial layout, emphasizing key elements like lane lines and building areas}.}
	\label{fig:combined}
	% \vspace{1.0cm}
\end{figure}

\subsection{Map Processing}
Since our method incorporates navigation map inputs and BEV semantic segmentation labels, it requires appropriate processing and fusion of different map data sources:
\subsubsection{Map Rasterization}
For accessibility and comprehensive coverage, we utilize OSM\cite{haklay2008openstreetmap} as our navigation map data source, as depicted in Fig. \ref{fig:left}. OSM represents buildings using polygonal areas, roads using polylines, and traffic signals and other points of interest (PoI) using nodes.
As shown in Fig. \ref{fig:right}, we retain only essential elements like buildings, roads, and PoIs (traffic signals, poles), whose spatial arrangement provides crucial geometric constraints for localization.
For each query, we retrieve a patch of rasterized navigation map centered around the initial vehicle localization coordinates.

\subsubsection{Segmentation Labels}
The BEV semantic segmentation labels come from two sources. The drivable area labels are obtained from HD map data such as nuScenes \cite{caesar2020nuscenes}. As a complementary source, building and PoI labels are derived from the navigation map, such as OSM \cite{mooney2017review}.

\subsection{BEV Module}
This module is designed to extract image features and project them into the BEV space to obtain BEV features.
The visual input can be a monocular front-view image or multiple surround-view images.
The more images used, the broader the perception range, resulting in improved localization accuracy.
An example of surround-view images is illustrated as shown in Fig. \ref{fig:sub3}.
We choose the simple yet effective LSS\cite{lss} architecture as the backbone.

We adopt the EfficientNet \cite{tan2019efficientnet} as the Perspective-View (PV) Encoder to extract the image features $\mathbf{{F}_{I}}$. 
%$\mathbf{{F}_{I}} \in {\mathbb{R}}^{{N}\times{C_{pv}}\times{H_{pv}}\times{W_{pv}}}$, where $N$ is the number of cameras, and $C_{pv}$, $H_{pv}$, $W_{pv}$ are feature channel, height and width separately.
Following the LSS procedure, we combine the extrinsic and intrinsic parameters to project $\mathbf{{F}_{I}}$ to BEV space of size ${{H}_{bev} \times {W}_{bev}}$.
We consider the longitudinal observation range to be wider than the lateral range, and thus set the spatial dimensions such that ${{H}_{bev} = 2{W}_{bev}}$.
In different up-sampling stages of the BEV Decoder module, we extract the low-resolution, high-channel coarse features $\mathbf{F_{bev}^{c}} \in {\mathbb{R} ^ {D_c \times \frac{1}{2}H_{bev} \times \frac{1}{2}W_{bev}}}$ and the high-resolution, low-channel fine features $\mathbf{F_{bev}^{f}} \in {\mathbb{R} ^ {D_f \times H_{bev} \times W_{bev}}}$, which are used for subsequent two-stage coarse-to-fine feature registration.
% In different up-sampling stages of the BEV Decoder, we extract coarse features $\mathbf{F_{bev}^{c}} \in {\mathbb{R} ^ {D_c \times \frac{1}{2}H_{bev} \times \frac{1}{2}W_{bev}}}$ with low resolution and high channel count, as well as fine features $\mathbf{F_{bev}^{f}} \in {\mathbb{R} ^ {D_f \times H_{bev} \times W_{bev}}}$ with high resolution and low channel count. These features are used for subsequent two-stage coarse-to-fine feature registration.
We supervise this module with a BEV semantic segmentation auxiliary task, 
which can better constrain the model's learning objective while also effectively improving localization accuracy.

\subsection{Map \revise{U-Net}}
We adopt a U-Net architecture to extract features from the rasterized maps.
To mitigate the modality gap between map features and visual BEV features, we innovatively introduce a BEV segmentation auxiliary task for this module. 
We employ a VGG-16\cite{simonyan2014very} backbone for encoding the map features. 
Similar to the BEV Module, at different stages of the Map Decoder, we also extract coarse and fine-level map features $\mathbf{F_{map}^{c}} \in \mathbb{R} ^ {D_c \times \frac{1}{2}{H_{map} \times \frac{1}{2}{W_{map}}}}$ and $\mathbf{F_{map}^{f}} \in \mathbb{R} ^ {D_f \times {H_{map} \times {W_{map}}}}$ for subsequent hierarchical feature registration. 
Here exists the relationship where $ H_{map} = H_{bev}$ and $ W_{map} = 2W_{bev}$, facilitating feature fusion.
We utilize the same BEV segmentation labels as the BEV module to supervise this module, constraining the disparity between the two types of features.

\subsection{Neural Localization Module} % wuhang
\label{sec:neural_loc}
%We adopt a Transformer decoder similar to DETR \cite{DETR} to compute cross-attention between BEV features and SD map features. 
This module is responsible for the fusion of map and visual features as well as the decoding of pose offsets, making it the core module of MapLocNet. We designed various architectures for pose decoders, and through extensive experiments, determined coarse-to-fine feature registration as the final optimal solution. Other comparative approaches will be discussed in the section \ref{sec:comparison}.

We formulate neural localization as %an image 
a feature
registration task. Inspired by C2F-ViT\cite{mok2022affine}, we adopt a transformer encoder to perform self-attention computation on the fused visual BEV and map features in a coarse-to-fine manner.
The modules for coarse and fine registration share the same architecture.
As the widths of BEV features and map features are different, we zero-pad the BEV features in the width dimension to match the width of the map features.
Considering the computing consuming, we downsampled the BEV and map features $4$ times along the height and width dimension. Following the C2F-ViT, we also employ a $7 \times 7$ convolutional kernel to fuse them along channel dimension and flattened the fused features into sequential tokens for self-attention encoding of pose hidden features.

Since this is a pose-related  task, positional encoding is crucial. We experimented with both learned and fixed positional encoding methods, eventually opting for sinusoidal positional encoding similar to that used in \cite{transformer}. 
However, we shifted the origin of the positional encoding coordinates to the center of the feature map. 
The positional encoding is injected into the fused features through element-wise addition.

In each Neural Localization Module, we design $N$ repeated transformer encoder layers, where in practice we set $N=3$. 
Following this, there is a Pose Head consisting of 3 layers of MLP for pose decoding. 
The $3$-DoF pose offset $\boldsymbol{\hat{\xi}_{c}}$ estimated in the coarse feature registration stage is applied to the fine BEV features $\mathbf{F_{bev}^{f}}$. 
Subsequently, the spatially transformed BEV features, along with the fine map features $\mathbf{F_{map}^{f}}$, undergo fine feature registration to further narrow the gap from the ground truth pose and obtain $\boldsymbol{\hat{\xi}_{f}}$. 
%Then, the corrected BEV features undergo fine feature registration with the fine map features $\mathbf{F_{map}^{f}}$ to further narrow the gap from the ground-truth pose, yielding $\boldsymbol{\hat{\xi}_{f}}$. 
The cumulative outputs from both stages collectively serve as the final pose offset estimation.

\begin{table*}[ht]
  \renewcommand\arraystretch{1.3}
  \centering
  \setlength{\tabcolsep}{1.8mm}{
  \begin{tabular}{lcccccccccccc}
    \toprule
    \multicolumn{1}{c}{\multirow{2.5}*{Approach}} &
    \multirow{2.5}*{\#Cams} &
    \multicolumn{4}{c}{\centering Position Recall@$Xm$ $\uparrow$} &
    \multicolumn{4}{c}{\centering Orientation Recall@$X^\circ$ $\uparrow$} &
    \multirow{2.5}*{\#Param.(M)} & 
    \multirow{2.5}*{GFLOPs $\downarrow$} & 
    \multirow{2.5}*{FPS $\uparrow$} \\
    \cmidrule(r){3-6} \cmidrule(r){7-10}
    & &
    \multicolumn{1}{c}{\centering $1m$} &
    \multicolumn{1}{c}{\centering $2m$} &
    \multicolumn{1}{c}{\centering $5m$} &
    \multicolumn{1}{c}{\centering $10m$} &
    \multicolumn{1}{c}{\centering $1^\circ$} & 
    \multicolumn{1}{c}{\centering $2^\circ$} &
    \multicolumn{1}{c}{\centering $5^\circ$} &
    \multicolumn{1}{c}{\centering $10^\circ$} \\
    \midrule
    OrienterNet\cite{sarlin2023orienternet} & $1$ & $5.83$ & $18.92$ & $52.83$ & $66.21$ & $32.13$ & $41.56$ & $65.63$ & $80.41$ & $54.9$ & $161.9$ & $8.1$ \\ % estimate version
    MapLocNet One-Stage (Ours) & $1$ & $\uline{7.14}$ & $\uline{22.39}$ & $\uline{60.48}$ & $\uline{83.90}$ & $\uline{35.68}$ & $\uline{60.40}$ & $\uline{87.49}$ & $\uline{95.31}$  & $20.7$  & $\textbf{49.81}$ & $\textbf{38.5}$ \\
    \textbf{MapLocNet (Ours)} & $1$ & $\textbf{8.96}$ & $\textbf{27.05}$ & $\textbf{64.57}$ & $\textbf{85.86}$ & $\textbf{40.36}$ & $\textbf{65.31}$ & $\textbf{89.66}$ & $\textbf{96.17}$  & $22.9$ & $\uline{52.15}$ & $\uline{23.8}$ \\
    \midrule
    U-BEV*\cite{camiletto2023u} & $6$ & $\uline{16.89}$ & $\uline{41.60}$ & $71.33$ & $83.46$ & $-$ & $-$ & $-$ & $-$ & $-$ & $-$ & $-$ \\
    MapLocNet DETR (Ours) & $6$ & $10.40$ & $29.76$ & $66.81$ & $86.81$ & $42.59$ & $68.05$ & $90.83$ & $96.63$  & $20.5$ & $55.40$ & $19.6$ \\
    MapLocNet CA (Ours) & $6$ & $13.36$ & $35.16$ & $70.54$ & $88.25$ & $48.15$ & $73.09$ & $92.66$ & $97.34$  & $20.8$ & $\uline{53.80}$ & $\uline{23.1}$ \\
    MapLocNet One-Stage (Ours) & $6$ & $\uline{16.32}$ & $\uline{40.56}$ & $\uline{74.27}$ & $\uline{89.69}$ & $\uline{53.71}$ & $\uline{78.13}$ & $\uline{94.49}$ & $\uline{98.05}$  & $20.7$ & $\textbf{53.14}$ & $\textbf{24.4}$ \\
    \textbf{MapLocNet (Ours)} & $6$ & $\textbf{20.10}$ & $\textbf{45.54}$ & $\textbf{77.70}$ & $\textbf{91.89}$ & $\textbf{58.61}$ & $\textbf{84.10}$ & $\textbf{96.23}$ & $\textbf{98.62}$  & $22.9$ & $56.15$ & $18.2$ \\
    \bottomrule
  \end{tabular}}
  \caption{Localization results on nuScenes dataset. * denotes that the method only predicts 2-DoF pose offset (data w/o orientation noise). The inference speed in FPS for all models was measured on an NVIDIA V100 GPU.}
  \label{tab:localization_result}
\end{table*}

\subsection{Loss Function}
During training, since we have the ground-truth localization $\mathbf{p}$, it is easier to simulate the noisy localization $\mathbf{\check{p}}$ by applying the transformation $\mathbf{T}$ to $\mathbf{p}$, where $\mathbf{T}$ is the inverse of the pose offset matrix from Eq. \ref{eq:1}, denoted as $\mathbf{T_{err\leftarrow GT}}$, and simplified to the 3-variable form $\boldsymbol{\xi_{1}}$.

As our method involves two stages, it can be considered as applying two successive transformations to the ground truth localization to obtain the noisy localization $\mathbf{\check{p}}$. Hence, the following relationship exists:
\begin{equation}
	 \mathbf{\hat{T}} =  \mathbf{\hat{T}_2} \cdot  \mathbf{\hat{T}_1}.
    \label{eq:2}
\end{equation}
$ \mathbf{\hat{T}_1}$ is the prediction from the first stage, representing the transformation $\mathbf{\hat{T}_{c\leftarrow GT}}$ from the ground truth localization to the coarse localization, corresponding to $\boldsymbol{\hat{\xi}_{c}}$.
$\mathbf{\hat{T}_2}$ represents the prediction from the second stage, representing the transformation $\mathbf{\hat{T}_{f\leftarrow c}}$  from the coarse estimated localization to the fine localization, corresponding to $\boldsymbol{\hat{\xi}_{f}}$.

Based on the relationships above, the coarse pose offset $\boldsymbol{\hat{\xi}_{c}}$ is supervised by ground truth $\boldsymbol{\xi_{1}}$, with the Smooth L1 Loss:
\begin{equation}
	\mathcal{L}_{c} = \left \|\boldsymbol{\hat{\xi}_{c}}, \boldsymbol{\xi_{1}} \right \|_{S1}.
    \label{eg:3}
\end{equation}
Where $ \left \| \cdot  \right \|_{S1} $ denotes the Smooth L1 Loss.
The ground truth $\mathbf{T_{2}}$ corresponding to $\mathbf{\hat{T}_{2}}$ relies on the first stage prediction $\mathbf{\hat{T}_{1}}$ and the ground truth $\mathbf{T}$, obtained from:
\begin{equation}
	\mathbf{T_{2}} = \mathbf{T} \cdot (\mathbf{\hat{T}_{1}})^{-1}.
    \label{eg:4}
\end{equation}
Denoting the simplified form of $\mathbf{T_{2}}$ as $\boldsymbol{\xi_{2}}$, it is used to supervise $\boldsymbol{\hat{\xi}_{f}}$ also with a Smooth L1 Loss:
\begin{equation}
	\mathcal{L}_{f} = \left \|\boldsymbol{\hat{\xi}_{f}}, \boldsymbol{\xi_{2}} \right \|_{S1}.
    \label{eg:5}
\end{equation}

To better supervise the BEV module, we designed a BEV segmentation auxiliary task, 
following the approach outlined in LSS \cite{lss},
to accelerate training convergence and improve model performance. We employ a Binary Cross-Entropy (BCE) loss to constrain the semantic segmentation output of the BEV module against the ground truth labels, resulting in $\mathcal{L}_{bev}$. On the other hand, to encourage the map features and visual BEV features to be as close as possible, thereby reducing the difficulty of feature registration, we also designed a BEV segmentation auxiliary task for the Map U-Net, with ground truth labels corresponding to map regions. Similarly, we adopted the BCE loss function to obtain $\mathcal{L}_{map}$.

% The overall loss is defined as follows, weighted by  ${\lambda}_{c}$,  ${\lambda}_{f}$, ${\lambda}_{bev}$ and ${\lambda}_{map}$:
% \begin{equation}
% \begin{split}
%     \mathcal{L} = {\lambda}_{c}\cdot{\mathcal{L}_{c}} + {\lambda}_{f}\cdot{\mathcal{L}_{f}} + {\lambda}_{bev}\cdot{\mathcal{L}_{bev}} + 
%     {\lambda}_{map}\cdot{\mathcal{L}_{map}}
% \end{split}
% \end{equation}
% In the training process, we set all weights to 1.

The overall loss function is defined as:
\begin{equation}
\begin{split}
    \mathcal{L} = {\lambda}_{c}\cdot{\mathcal{L}_{c}} + {\lambda}_{f}\cdot{\mathcal{L}_{f}} + {\lambda}_{bev}\cdot{\mathcal{L}_{bev}} + 
    {\lambda}_{map}\cdot{\mathcal{L}_{map}}
\end{split}
\end{equation}
where ${\lambda}_{c}$,  ${\lambda}_{f}$, ${\lambda}_{bev}$ and ${\lambda}_{map}$ are weighting factors. During training, all weights are set to 1.
 % wuhang

\section{Experiments} % wuhang

\begin{table*}[ht]
  \renewcommand\arraystretch{1.2}
  \centering
  \begin{tabular}{ccccccccccccccc} 
    \toprule
    \multirow{2.5}*{Lanes} & 
    \multirow{2.5}*{Buildings} & 
    \multirow{2.5}*{Nodes} &
    \multicolumn{4}{c}{\centering Longitudinal Recall@$Xm$ $\uparrow$} &
    \multicolumn{4}{c}{\centering Lateral Recall@$Xm$ $\uparrow$} &
    \multicolumn{4}{c}{\centering Orientation Recall@$X^\circ$ $\uparrow$} \\
    \cmidrule(r){4-7} \cmidrule(r){8-11} \cmidrule(r){12-15}
    & & &
    \multicolumn{1}{c}{\centering $1m$} &
    \multicolumn{1}{c}{\centering $2m$} &
    \multicolumn{1}{c}{\centering $5m$} &
    \multicolumn{1}{c}{\centering $10m$} &
    \multicolumn{1}{c}{\centering $1m$} &
    \multicolumn{1}{c}{\centering $2m$} &
    \multicolumn{1}{c}{\centering $5m$} &
    \multicolumn{1}{c}{\centering $10m$} &
    \multicolumn{1}{c}{\centering $1^\circ$} & 
    \multicolumn{1}{c}{\centering $2^\circ$} &
    \multicolumn{1}{c}{\centering $5^\circ$} & 
    \multicolumn{1}{c}{\centering $10^\circ$} \\
    \midrule
    $\checkmark$ & $\checkmark$ & $\checkmark$ & $32.08$ & $54.20$ & $81.97$ & $93.34$ & $48.71$ & $71.91$ & $90.66$ & $97.11$ & $58.61$ & $84.10$ & $96.23$ & $98.62$ \\
    $\checkmark$ & $\checkmark$ & $ $ & $28.07$ & $48.75$ & $77.64$ & $90.79$ & $41.48$ & $64.73$ & $88.00$ & $96.54$ & $51.44$ & $75.74$ & $93.61$ & $97.67$ \\
    $\checkmark$ & $ $ & $ $ & $22.92$ & $41.53$ & $72.58$ & $88.80$ & $34.66$ & $58.12$ & $85.72$ & $96.04$ & $42.59$ & $68.05$ & $90.83$ & $96.63$ \\
    \bottomrule
  \end{tabular}
  \caption{Localization results of various combination of input map elements on nuScenes dataset.}
  \label{tab:ablation_input_map_ele}
\end{table*}

\begin{table*}[ht]
  \renewcommand\arraystretch{1.2}
  \centering
  \setlength{\tabcolsep}{1.8mm}{
  \begin{tabular}{ccccccccccccccc} 
    \toprule
    \multirow{2.5}*{Pose Loss} &
    \multirow{2.5}*{BEV Loss} &
    \multirow{2.5}*{Map Loss} &
    \multicolumn{4}{c}{\centering Longitudinal Recall@$Xm$ $\uparrow$} &
    \multicolumn{4}{c}{\centering Lateral Recall@$Xm$ $\uparrow$} &
    \multicolumn{4}{c}{\centering Orientation Recall@$X^\circ$ $\uparrow$} \\
    \cmidrule(r){4-7} \cmidrule(r){8-11} \cmidrule(r){12-15}
    & & &
    \multicolumn{1}{c}{\centering $1m$} &
    \multicolumn{1}{c}{\centering $2m$} &
    \multicolumn{1}{c}{\centering $5m$} &
    \multicolumn{1}{c}{\centering $10m$} &
    \multicolumn{1}{c}{\centering $1m$} &
    \multicolumn{1}{c}{\centering $2m$} &
    \multicolumn{1}{c}{\centering $5m$} &
    \multicolumn{1}{c}{\centering $10m$} &
    \multicolumn{1}{c}{\centering $1^\circ$} & 
    \multicolumn{1}{c}{\centering $2^\circ$} &
    \multicolumn{1}{c}{\centering $5^\circ$} &
    \multicolumn{1}{c}{\centering $10^\circ$} \\
    \midrule
    $\checkmark$ & $\checkmark$ & $\checkmark$ & $32.08$ & $54.20$ & $81.97$ & $93.34$ & $48.71$ & $71.91$ & $90.66$ & $97.11$ & $58.61$ & $84.10$ & $96.23$ & $98.62$ \\
    $\checkmark$ & $\checkmark$ & $ $ & $27.21$ & $47.62$ & $76.89$ & $90.49$ & $40.48$ & $63.73$ & $87.69$ & $96.52$ & $50.01$ & $74.54$ & $93.20$ & $97.53$ \\
    $\checkmark$ & $ $ & $ $ & $21.31$ & $39.21$ & $70.56$ & $87.95$ & $32.32$ & $55.62$ & $84.85$ & $95.83$ & $40.36$ & $65.31$ & $89.66$ & $96.17$ \\
    \bottomrule
  \end{tabular}}
  \caption{Localization results of various combination of loss functions on nuScenes dataset.}
  \label{tab:ablation_loss_func}
\end{table*}

\subsection{Experimental Settings}

\subsubsection{Dataset}
Our proposed approach was trained and validated using two autonomous driving datasets, nuScenes\cite{caesar2020nuscenes} and Argoverse\cite{chang2019argoverse}, to ensure a comprehensive evaluation. The nuScenes dataset comprises $1000$ driving sequences captured in Boston and Singapore. We used the default train split of nuScenes, consisting of $850$ sequences. The nuScenes validation set, comprising $150$ sequences, served as our evaluation benchmark. The Argoverse dataset, featuring $113$ scenes recorded in Miami and Pittsburgh, was utilized, with $65$ scenes allocated to the training split and $24$ scenes to the validation split.
To address the absence of navigation map data in both nuScenes and Argoverse datasets, we enriched our dataset by acquiring navigation maps from OSM for the corresponding geographic regions. 
We aligned the navigation map with the HD map through localization coordinate transformation, following the methodology outlined in BLOS-BEV \cite{blos_bev}.
Fig. \ref{fig:combined} visualizes the alignment of local section of the navigation map information and a frame from the nuScenes dataset at the same location.

\subsubsection{Implementation Details} %wuhang
\paragraph{Network Settings}
The MapLocNet employs 6 surround-view images as the visual input unless otherwise specified. We utilize the EfficientNet-B0\cite{tan2019efficientnet} architecture as the image backbone, and all input images are resized to $128\times352$ resolution. During the training stage, we apply essential image data augmentations to enhance model robustness including random cropping, random flipping, and the one random drop of camera input. 
% The ego vehicle perceptron range of BEV space is defined as [$-64m$, $64m$, $0.5mpp$] for the longitudinal-axis, [$-32m$, $32m$, $0.5mpp$] for the lateral-axis, and the bins range of depth distribution is [$4m$, $60m$, $1mpp$].
The ego vehicle's perceptual range in BEV space is defined as [$-64m$, $64m$] along the longitudinal axis and [$-32m$, $32m$] along the lateral axis, both with a resolution of 0.5 meters per pixel ($mpp$). 
The depth distribution is binned within the range [$4m$, $60m$] at a resolution of $1mpp$.
% \paragraph{Navigation Map Settings} 
For each frame, we take a $128m\times128m$ patch from rasterized navigation map centered on the ego vehicle's position, with a resolution of $0.5mpp$.

\paragraph{Simulating 3-DoF GPS Error}
We first align the pose and scale of rasterized navigation map with the HD map from nuScenes and Argoverse datasets. During training, we take a patch %slightly larger than $128\times128m$ 
from the rasterized navigation map, centered on the ego vehicle's position. To simulate GPS errors, we apply random rotations $\theta \in [-30^\circ, 30^\circ]$ and translations $\mathbf{t} \in [-30m, 30m]$ to this patch. We then crop the central  $128m\times128m$ area as the biased map input of the MapLocNet.

\paragraph{Training Details}
We train the model using $8$ NVIDIA V100 GPUs for $200$ epochs, which takes around \revise{48} hours to converge. The model is optimized using an AdamW optimizer\cite{Loshchilov2017FixingWD} with a weight decay of 1e-7, batch size of \revise{8}, and initial learning rate of \revise{1e-4}. We use a cosine annealing scheduler to adjust the learning rate during training.

\subsection{Localization Results and Comparison}  %siyuan  
\label{sec:comparison}
\subsubsection{Comparison Methods}
\paragraph{OrienterNet}
We employed the official implementation of OrienterNet \cite{sarlin2023orienternet} for training and evaluation on both nuScenes and Argoverse datasets. To ensure a fair comparison, given OrienterNet's limitation to monocular input, we conducted a parallel set of experiments using single-camera input for our method.

\paragraph{U-BEV}
Since U-BEV's task is similar to ours, we directly referenced the data provided in its paper\cite{camiletto2023u}. It's worth noting that its localization results do not include orientation predictions.
Consequently, the initial localization may lack heading angle error perturbations, which simplifies the task to some extent. 
Given its utilization of 6 surround-view images, we categorize it within the reference group for 6-camera configurations.

\paragraph{MapLocNet DETR}
We take inspiration from the decoder design in DETR\cite{DETR}, creatively treating the pose offset as the query $\mathbf{Q}$ to retrieve the fused features from the visual BEV features and map features. The feature fusion follows the approach described in section \ref{sec:neural_loc}. The features processed by the DETR decoder are then passed through the same 3-layer MLP Pose Head for pose decoding.

\paragraph{MapLocNet CA}
Inspired by LoFTR\cite{sun2021loftr} and GeoTransformer\cite{qin2022geometric}, we designed our neural localization module using a Cross-Attention (CA) module. We treat visual features as queries $\mathbf{Q}$ and map features as key $\mathbf{K}$ and value $\mathbf{V}$, %and vice versa, 
enabling cross-domain attention computation. The resulting features are then decoded by the same Pose Head for pose estimation.

\paragraph{MapLocNet One-Stage}
Our method is hierarchical, which to some extent affects inference speed. We want to investigate whether our method can meet usage requirements with only one stage, namely coarse feature registration, under constrained computational resources. Therefore, we tested a one-stage version of MapLocNet here. 
To minimize computational complexity, we employed coarse features instead of fine features for our one-stage experiments.
The sole difference from the hierarchical version is that we omitted fine feature registration and directly used the output of the initial coarse stage as the final result.
We expect the one-stage version to balance localization accuracy and inference speed, while the coarse-to-fine version can push the upper bound of localization accuracy.

\subsubsection{\revise{Localization Results}}
\paragraph{nuScenes}
For simplicity, we collectively term both one-stage and coarse-to-fine approaches as feature registration (FR) architecture.
As shown in Tab. \ref{tab:localization_result}, among the experimental groups with 6 cameras, the coarse-to-fine FR architecture achieved the best localization performance. 
Our one-stage FR architecture demonstrated the highest efficiency, achieving 24.4 frames per second (FPS). 
In the monocular experimental group, we compared the localization performance of the FR architecture with OrienterNet. 
Our approach surpassed OrienterNet in both computational efficiency and accuracy, notably exceeding OrienterNet's speed by approximately 30 FPS.

\begin{table}[h]
  \renewcommand\arraystretch{1.2}
  \centering
   \setlength{\tabcolsep}{1.1mm}{
  \begin{tabular}{ccccccccc}
    \toprule
    \multirow{2.5}*{Approach} &
    \multicolumn{4}{c}{\centering Position Recall@$Xm$ $\uparrow$} &
    \multicolumn{4}{c}{\centering Orientation Recall@$X^\circ$ $\uparrow$} \\
    \cmidrule(r){2-5} \cmidrule(r){6-9}
    & 
    \multicolumn{1}{c}{\centering $1m$} &
    \multicolumn{1}{c}{\centering $2m$} &
    \multicolumn{1}{c}{\centering $5m$} &
    \multicolumn{1}{c}{\centering $10m$} &
    \multicolumn{1}{c}{\centering $1^\circ$} & 
    \multicolumn{1}{c}{\centering $2^\circ$} &
    \multicolumn{1}{c}{\centering $5^\circ$} &
    \multicolumn{1}{c}{\centering $10^\circ$} \\
    \midrule
    O-1 & $8.56$ & $21.20$ & $54.90$ & $72.16$ & $18.72$ & $31.07$ & $63.05$ & $81.97$ \\
    M-1 & $\uline{9.12}$ & $\uline{27.61}$ & $\uline{66.31}$ & $\uline{88.71}$ & $\uline{41.22}$ & $\uline{66.54}$ & $\uline{90.32}$ & $\uline{96.53}$ \\
    M-6 & $\textbf{23.26}$ & $\textbf{47.24}$ & $\textbf{79.13}$ & $\textbf{94.33}$ & $\textbf{62.35}$ & $\textbf{86.28}$ &$\textbf{96.24}$ & $\textbf{98.61}$ \\
    \bottomrule
  \end{tabular}}
  \caption{Localization results on Argoverse dataset. Specifically, approach \textit{O-1} represents OrienterNet\cite{sarlin2023orienternet} with single-view input, while \textit{M-1} and \textit{M-6} denote MapLocNet with single-view and six-view inputs, respectively.}
  \label{tab:Argoverse_result}
\end{table}

\paragraph{Argoverse}
To further demonstrate the capabilities of our model, we conducted experiments on the Argoverse \cite{chang2019argoverse} dataset. Leveraging pre-trained weights obtained from the nuScenes dataset, we fine-tuned our model on the Argoverse dataset. 
We also compared our approach with OrienterNet, employing identical training strategies.
Remarkably, as depicted in Tab. \ref{tab:Argoverse_result}, our model exhibits superior  localization performance across both single-view camera and surround-view camera configurations.
% Our method significantly outperforms OrienterNet in localization accuracy, not only with single-camera input but also with multi-camera surround-view input, demonstrating the robustness and versatility of our approach.
It significantly outperforms OrienterNet in accuracy across all input settings, highlighting the robustness and versatility of our approach.

\subsubsection{Results Visualization}
\begin{figure}[t]
	\centering
    % \vspace{0.5cm}
	\includegraphics[width=0.48\textwidth]{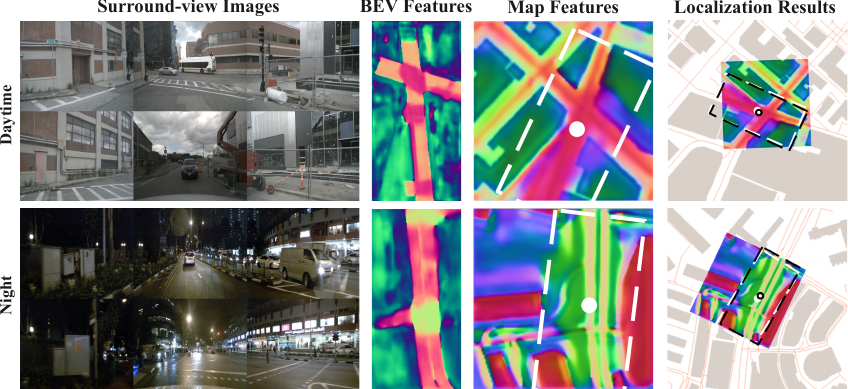}
	\caption{Visualization of localization results in the nuScenes dataset during day and night. The middle two columns depict high-resolution, low-channel BEV features and map features, respectively. The white dots and bounding boxes in the map features represent the GT locations and orientations of the BEV features. In the last column, black dots and bounding boxes represent the corrected locations and orientations on the map after applying the offsets predicted by the model.
 }
	\label{visuliazation of results}
	%\vspace{-1.0cm}
\end{figure}

To illustrate the model's performance intuitively, Fig. \ref{visuliazation of results} displays only the high-resolution, low-channel BEV features and map features utilized in the second-stage registration. The high-dimensional features employed in the initial coarse registration are omitted due to their visual complexity. Our experiments reveal a minor performance degradation in night scenes, attributed to reduced building visibility. Nevertheless, the model exhibits robust localization capabilities across both daytime and nighttime conditions.

\subsection{Ablation Study}
We performed comprehensive ablation studies to evaluate the influence of various OSM element combinations and loss function configurations on model performance.

\subsubsection{Input OSM Elements}
We conducted an ablation study on three key map elements: lanes, buildings, and nodes (including traffic lights and signs). Given their environmental prevalence, we sequentially removed nodes and buildings from the input.
% As shown in Tab. \ref{tab:ablation_input_map_ele}, the gradual degradation of localization performance suggest that all three selected elements positively contribute to the model's localization performance. The decrease in localization performance is larger when removing buildings compared to removing nodes. This implies that buildings have a greater impact on localization than nodes. Even with only lanes as input, the model retains a significant portion of its localization performance, indicating that lanes play the most crucial role in localization.
Tab. \ref{tab:ablation_input_map_ele} demonstrates that all three elements contribute positively to localization performance. The removal of buildings caused a more significant performance drop compared to nodes, indicating their greater impact on localization. Notably, the model maintained substantial performance with lanes alone, suggesting their crucial role in localization.
We posit that from a BEV perspective, the learning complexity decreases from nodes to buildings to lanes, while their environmental prevalence increases. This correlation aligns with their increasing importance in localization performance.

\subsubsection{Loss Functions}

This experiment investigates the impact of auxiliary segmentation tasks on localization performance.
We introduce the BEV Loss, which guides feature learning in the visual branch, and the Map Loss, which uses the same semantic labels to reduce feature modality differences between visual and map branches.
As shown in Tab. \ref{tab:ablation_loss_func}, %the addition of visual BEV segmentation loss supervision significantly improves the model's localization performance. 
incorporating visual BEV segmentation loss supervision substantially enhances the model's localization performance.
We posit that this loss improves the model's comprehension of environmental structures, providing clearer localization cues. 
After adding map segmentation loss supervision, the model's localization performance is further improved. 
We hypothesize that a modality gap exists between the rasterized map and visual BEV representations.
By unifying semantic supervision across both branches, we mitigate this modality discrepancy, thereby improving the model's localization capabilities.

\section{Conclusion} %qintong
In this work, we propose MapLocNet, a novel approach that achieves highly accurate and reliable localization by integrating surround-view images and navigation maps. Our method effectively addresses the challenge of re-localization under significant  positioning drift, particularly in complex urban environments. % such as urban canyons and elevated bridges. 
Through hierarchical feature registration, our approach surpasses existing methods in both localization accuracy and inference speed on both the nuScenes and Argoverse datasets. 
The two-stage feature registration ensures high precision, while the one-stage approach balances accuracy and speed. 
Supervised BEV segmentation in both BEV and map modules mitigates modality gap, enhancing localization capability. 
Our method represents a notable advancement towards robust and precise vehicle localization, facilitating safe and reliable autonomous driving.

\textbf{Limitations and Future Work:} 
Our method currently uses single-frame inference and relies on BEV semantic supervision. Future work will focus on incorporating multi-frame information, exploring vectorized navigation map representations, and developing methods to mitigate multi-modal discrepancies without semantic supervision. These advancements aim to enhance localization accuracy, reduce computational overhead, and improve overall performance.

\clearpage

\bibliography{reference.bib}

\end{document}